# Generalizing Jeffrey Conditionalization

Carl G. Wagner
Mathematics Department
University of Tennessee
Knoxville, TN 37996

## Abstract

Jeffrey's rule has been generalized by Wagner to the case in which new evidence bounds the possible revisions of a prior probability below by a Dempsterian lower probability. Classical probability kinematics arises within this generalization as the special case in which the evidentiary focal elements of the bounding lower probability are pairwise disjoint. We discuss a twofold extension of this generalization, first allowing the lower bound to be any two-monotone capacity and then allowing the prior to be a lower envelope.

## 1 INTRODUCTION

The revision of prior $p$ to posterior $q = p(|E)$ is appropriate if and only if one judges that $q(E) = 1$ and that $q(A|E) = p(A|E)$ for all events $A$. *Radical probabilism* (Jeffrey, 1985) recoils from dogmatic judgments like $q(E) = 1$, but is fortunately not deprived thereby of a principled method of probability revision, employing instead of simple conditioning the generalization (Jeffrey, 1965),

$$q(A) = \sum_{E \in \mathcal{E}} \mu_E p(A|E). \qquad (1)$$

Here $\mathcal{E}$ is a countable collection of nonempty, pairwise disjoint events and $\{\mu_E : E \in \mathcal{E}\}$ a family of positive reals summing to one. To be justified in revising $p$ by (1) one must judge, based on the total evidence, that

$$q(E) = \mu_E, \quad \text{for every } E \in \mathcal{E}, \qquad (2)$$

and that, for all events $A$,

$$q(A|E) = p(A|E), \quad \text{for every } E \in \mathcal{E}. \qquad (3)$$

Jeffrey describes $q$ as coming from $p$ by *probability kinematics*, the analogy with mechanics residing in the "conservation of conditional probabilities" posited by (3).

It has been observed that formula (1) can be derived by Dempster's rule (Shafer, 1981), and by relative information minimization (May, 1976). More recently, three different asymmetrical rules for combining belief functions have been proposed (Ichihashi and Tanaka, 1991), each of which contains (1) as a special case. As we argue below in more detail, these approaches to the updating of assessments of uncertainty are seriously incomplete, furnishing only a generalization of (1), but no generalization of the key criterion (3) for implementing (1). What they provide are merely formal generalizations of Jeffrey's rule, or what have been called "mechanical updating" methods (Diaconis and Zabell, 1982). We would maintain that any updating formula true to the spirit of classical probability kinematics must be furnished with a warranting conservation-of-conditional probability criterion of some sort. In what follows, we outline such a generalization of Jeffrey conditionalization.

## 2 GENERALIZED PROBABILITY KINEMATICS

In the following generalization of classical probability kinematics, new evidence, rather than fixing certain values of the posterior $q$, merely places a lower bound on the values of $q$. We first examine the case in which the set function furnishing this lower bound is a *Dempsterian lower probability*. Full details may be found in Wagner (1990).

Let the finite set $X$ be equipped with a prior probability measure $p$, assumed to be positive on all nonempty subsets of $X$. Suppose that additional evidence, in conjunction with our prior evidence, enables us to assess a positive probability measure $u$ over subsets of a finite, related set of possibilities $Y$, and that our understanding of the relation between outcomes in $Y$ and those in $X$ is summarized in a function $\Gamma : Y \to 2^X - \{\emptyset\}$, where for each $y \in Y$, $\Gamma(y)$ denotes the set of outcomes in $X$ compatible with $y$.

As noted by Dempster (1967), $u$ and $\Gamma$ induce three interesting set functions $m$, $b$, and $a$, on $X$, defined for all $E \subseteq X$ by

$$m(E) = u\{y \in Y : \Gamma(y) = E\}, \qquad (4)$$

$$b(E) = \sum_{H \subseteq E} m(H) = u\{y \in Y : \Gamma(y) \subseteq E\} \quad (5)$$

and

$$a(E) = 1 - b(\overline{E}) = u\{y \in Y : \Gamma(y) \cap E \neq \emptyset\}. \quad (6)$$

Members of the family $\mathcal{E} = \{E \subseteq X : m(E) > 0\}$ are called *evidentiary focal elements*. Members of $\mathcal{E}$ are not in general pairwise disjoint.

Since $b(E)$ (respectively, $a(E)$) is the sum of the probabilities of all outcomes in $Y$ that entail (respectively, do not preclude) the event $E$, it is clear that the evidence manifested in $u$ and $\Gamma$ restricts possible revisions of $p$ to those probability measures bounded below by $b$ and above by $a$, the latter restriction being redundant in virtue of (6).

The probability measures on $X$ bounded below by $b$ are shown in Wagner (1990) to be precisely the marginalizations to $X$ of joint probability measures $Q$ on $X \times Y$ that are *compatible* with $u$ and $\Gamma$ in the sense that 1) the marginalization of $Q$ to $Y$ is $u$ and 2) $x \notin \Gamma(y) \Rightarrow Q(x,y) = 0$. Indeed, $b$ is the lower envelope of all such marginalizations. We may of course never arrive at a fully specified probability measure $Q$ on $X \times Y$. But we might judge, nevertheless, that were we to arrive at such a $Q$, it would satisfy, for all $A \subseteq X$ and all $E \in \mathcal{E}$,

$$Q(\text{``}A\text{''}|\text{``}E_*\text{''}) = p(A|E), \quad (7)$$

where $E_* = \{y \in Y : \Gamma(y) = E\}$, "$A$" $= A \times Y$, and "$E_*$" $= X \times E_*$. To adopt (7) is to judge that the total impact of the occurrence of the event $E_*$ is to preclude the occurrence of any outcomes $x \notin E$, and that, within $E$, $p$ can be assumed to remain operative in the assessment of relative uncertainties.

There may well be an infinite number of joint probability measures $Q$ compatible with $u$ and $\Gamma$, and satisfying the conservation-of-conditional-probability condition (7). Their marginalizations to $X$ are, however, identically equal to the probability measure $q$, defined for all $A \subseteq X$ by

$$q(A) = \sum_{E \in \mathcal{E}} m(E) p(A|E). \quad (8)$$

The probability $q$ is thus the uniquely acceptable revision of $p$ that is bounded below by $b$, given that (7) is judged to hold. This account furnishes a complete generalization of Jeffrey conditionalization, which amounts to the special case of the above in which the family $\mathcal{E}$ is pairwise disjoint. For in that case the conditions $q \geq b$ and (7) are equivalent to Jeffrey's criteria (2) and (3), with $\mu_E = m(E) = b(E)$, and (8) reduces to the classical kinematical rule (1).

## 3  BOUNDING POSTERIORS BY TWO-MONOTONE CAPACITIES

Dempsterian lower probabilities, arising from the projection of probability measures via compatibility relations, are highly structured set functions. Indeed, it can be shown for any $b$ defined by (5) that, for all $r \geq 2$ and every sequence $A_1, \ldots, A_r$ of subsets of $X$,

$$b(A_1 \cup \ldots \cup A_r) \geq \sum_{\substack{I \subseteq \{1,\ldots,r\} \\ I \neq \emptyset}} (-1)^{|I|-1} b(\cap_{i \in I} A_i). \quad (9)$$

Thus every Dempsterian lower probability is what Choquet (1953) calls an *infinitely monotone capacity*, or what Shafer (1976) calls a *belief function*.

While there is a substantial difference of opinion about what properties a set function must possess to qualify as a lower probability, no one would maintain that the infinitely monotone capacities exhaust the class of lower probabilities. Indeed, following Walley's (1981) adaptation of di Finetti (1974), we might define the lower probability of an event $A$ to be a number $c(A) \in [0, 1]$ such that we are prepared to pay any price less than $c(A)$ to receive one unit (of some appropriate good - perhaps Smith's (1961) "probability currency") if $A$ occurs and nothing otherwise. As Walley has shown, coherence of the values $c(A)$ puts rather mild structural restrictions on $c$, amounting only to the requirement that $c$ be the lower envelope of the set of all probability measures that dominate it. In particular, a coherent lower probability $c$ will always be *superadditive* ($A \cap B = \emptyset \Rightarrow c(A \cup B) \geq c(A) + c(B)$) and hence *monotone* $A \subseteq B \Rightarrow c(A) \leq c(B)$).

Suppose, generalizing the case treated in §2 above, that having assessed a positive prior $p$ on subsets of the finite set $X$, we are apprised of additional evidence that bounds any acceptable revision of $p$ below by the coherent lower probability $c$. Define the *Möbius transform*, $m$, of $c$ by

$$m(E) = \sum_{H \subseteq E} (-1)^{|E-H|} c(H). \quad (10)$$

It is easy to show that $m(\emptyset) = 0$ and that, for all $A \subseteq X$,

$$c(A) = \sum_{E \subseteq A} m(E). \quad (11)$$

In particular, $\sum_{E \subseteq X} m(E) = c(X) = 1$. When $c$ is a Dempsterian lower probability $b$, $m$, as defined by (10), is identical with $m$, as defined by (4). Along with (11), this demonstrates fairly conclusively that (10) is the right generalization of (4). In this setting, $m$ may take negative values (indeed, will take at least one negative value if $c$ is not infinitely monotone - see Chateauneuf and Jaffray (1989)). But, emboldened by our generalization of $m$, let us simply appropriate the carefully derived revision formula (8) from §2, and see what it can do for us here, i.e., let us consider the set function $q$, defined for all $A \subseteq X$ by

$$q(A) = \sum_{E \in \mathcal{E}} m(E) p(A|E) \quad (12)$$

where $m$ is now defined by (10) and $\mathcal{E} = \{E \subseteq X : m(E) \neq 0\}$.

Notwithstanding the fact that some of the numbers $m(E)$ in (12) may be negative, $q$ is *always* a probability measure, as a consequence of the monotonicity of $c$. More strikingly, if $c$ is 2-monotone ($c(A \cap B) \geq c(A) + c(B) - c(A \cap B)$), then $q$ *always* dominates $c$. Indeed, these results characterize monotonicity and 2-monotonicity. Proofs of these theorems may be found in Sundberg and Wagner (1990).

So if additional evidence places a 2-monotone lower bound $c$ on possible revisions of the prior $p$, the probability measure $q$ defined by (12) is at least in the running to be chosen as the posterior. But we do not yet have a criterion, of the type furnished by (7) when $c$ is a Dempsterian lower probability, that would single out $q$ as the uniquely acceptable posterior. Hence, at this point, (12) has only the status of a formal generalization of (8) to the case of a 2-monotone lower bound.

## 4  REVISING A PRIOR LOWER PROBABILITY

Suppose that, having assessed a coherent lower probability $\ell$ over subsets of $X$, we are apprised of additional evidence establishing with certainty that the true state of affairs lies in the subset $E$, so that any revision $\lambda$ of $\ell$ must satisfy $\lambda(E) = 1$. A natural way to extend $\lambda$ to arbitrary subsets $A$ is to set

$$\lambda(A) = \tag{13}$$
$$\ell^{(b)}(A|E) := \inf\{p(A|E) : p \text{ is a probability}$$
$$\text{measure dominating } \ell \text{ and } p(E) > 0\}.$$

This revision method, known as *Bayesian conditioning*, goes back at least as far as Dempster (1967). It can be applied as long as $\ell(\overline{E}) < 1$, even if $\ell(E) = 0$. If $\ell$ is 2-monotone, one can establish the nice formula

$$\ell^{(b)}(A|E) = \frac{\ell(A \cap E)}{\ell(A \cap E) + 1 - \ell(A \cup \overline{E})} \tag{14}$$

A proof of (14) appears in Sundberg and Wagner (1991), *where it is also shown that if $\ell$ is $r$-monotone (i.e., satisfies (9) with $b$ replaced by $\ell$ for the fixed integer $r$), then so is $\ell^{(b)}(\cdot|E)$*. This generalizes earlier partial results of Walley (1981), Jaffray (1990), and Fagin and Halpern (1990). Preservation of $r$-monotonicity under Bayesian conditioning puts this type of conditioning formally on par with two other types, *geometric conditioning* (Suppes and Zanotti, 1977) defined by

$$\ell^{(g)}(A|E) := \frac{\ell(A \cap E)}{\ell(E)} \tag{15}$$

and *Dempsterian conditioning* (Dempster, 1967), defined by

$$\ell^{(d)}(A|E) := \frac{\ell(A \cup \overline{E}) - \ell(\overline{E})}{1 - \ell(\overline{E})} \tag{16}$$

Formulae (14), (15), and (16) agree if $\ell(E) + \ell(\overline{E}) = 1$, but in general they disagree, it being certain only that $\ell^{(b)}(A|E)$ exceeds neither $\ell^{(g)}(A|E)$ nor $\ell^{(d)}(A|E)$. Bayesian conditioning is thus the most conservative of these methods. Here again, however, results are purely formal. The difficult work of articulating criteria for employing these conditioning methods remains to be done.

Clearly one encounters challenging problems in revising a prior lower probability even in the "dogmatic" case where evidence renders it certain that the truth lies in $E$. In the spirit of radical probabilism, however, one ought to investigate the problem of revising a prior $\ell_1$, given evidence that simply places a lower bound $\ell_2$ on possible revisions $\lambda$.

One revision formula worth exploring is

$$\lambda(A) = \tag{17}$$
$$\inf\{\sum_{E \in \mathcal{E}_2} m_2(E)p(A|E) : p \text{ is a probability}$$
$$\text{measure dominating } \ell_1 \text{ and positive on } \mathcal{E}_2\},$$

where $m_2$ is the Möbius transform of $\ell_2$ and $\mathcal{E}_2$ its set of focal elements. A theorem of Sundberg and Wagner (1990) mentioned earlier guarantees that $\lambda$, when well defined, is a coherent lower probability dominating $\ell_2$, as long as $\ell_1$ is coherent and $\ell_2$ is 2-monotone. Note that (12) and (13) are special cases of (17). When $\ell_2$ is a Dempsterian lower probability there is a criterion in the spirit of (7) for revising $\ell_1$ by (17). Whether such a criterion can be articulated in more general cases remains to be seen.

Note that (17) is not in computationally tractable form (neither was (13) until (14) came to light), although there are easily computable lower bounds on $\lambda$. If $\lambda$ is sometimes the proper revision of $\ell_1$, and if no simple formula for $\lambda$ emerges, then we may need to adjust to the idea of merely approximating an ideal posterior. Of course there may be superior alternatives to (17) in special cases, just as (15) and (16) may be on occasion superior to (13). A thorough investigation of this issue would appear to be both mathematically and philosophically interesting.

## 5  OTHER APPROACHES

The generalizations of classical probability kinematics described above furnish a perspective on attempted deconstructions of Jeffrey's rule as uninteresting special cases of (1) relative information minimization and (2) Dempster's rule, and suggest that these attempts are fundamentally misguided.

In the first case it has been observed (e.g., by May (1976)) that Jeffrey's formula (1), rather than being derived from (2) and (3), can be derived by minimizing the relative information measure

$$I(q,p) := \sum_x q(x) \log(q(x)/p(x)) \tag{18}$$

over those $q$ simply satisfying (2). Relative information minimization methods (also called maxent methods) may appear to furnish a powerful general method of updating a prior, subject to any constraints that yield a closed convex subset of possible posterior probability measures. But some convincing arguments have been advanced against maxent (e.g., by Skyrms (1987)) as a general updating method.

More particularly, when maxent is used to supplant Jeffrey's rule, the key conservation-of-conditional-probability criterion (3) for implementing that rule is discarded, resulting in what Diaconis and Zabell (1982) label "mechanical updating." But the decisive argument against viewing Jeffrey's rule as a special case of maxent is that the natural generalization of that rule furnished by (8) need *not* minimize $I(q,p)$ among all $q \geq b$ (Wagner, 1990). Since (8) is demonstrably the correct way to update $p$, given (7) and the restriction $q \geq b$, this furnishes a counterexample to the general reasonableness of maxent updating, and also shows that it is simply a fluke that maxent and Jeffrey's rule coincide.

As for using Dempster's rule in place of Jeffrey's, Shafer (1981) has shown how to construct a belief function $\beta$ such that $p \oplus \beta$, the result of combining the prior $p$ with $\beta$ by Dempster's rule, coincides with $q$, as defined by (1). Indeed, one can cook up a number of different belief functions $\beta$ with this property, which suggests a certain artificiality of construction. As in the case of maxent, the key conservation condition (3) is obscured by this analysis. Interestingly, for the generalization (8) of (1), one can also construct a belief function $\beta$ (not, by the way, the naturally occuring lower bound $b$) such that $p \oplus \beta = q$ (Wagner, 1990). But when one enters the realm of probability revision subject to a merely two-monotone lower bound one is (probably in the case of (12) and definitely in the case of (17)) outside the area of application of Dempster's rule, which applies only to belief functions and always yields a belief function. Thus a sufficiently broad generalization of probability kinematics can be expected to transcend even a formal rendering in terms of Dempster's rule.

We conclude by considering three asymmetrical rules for combining belief functions, each of which formally generalizes (1), and indeed (8). Given belief functions $b_1$ and $b_2$ on $X$ with associated Möbius transforms (or basic probability assignments) $m_1$, and $m_2$, Ichihashi and Tanaka (1989) define (though with different notation)

$$m_1|m_2(H) = \sum_{\substack{(E,F) \in \mathcal{E}_1 \times \mathcal{E}_2 \\ E \cap F = H}} \frac{m_1(E)m_2(F)}{1 - b_1(\overline{F})} \quad (19)$$

$$m_1\|m_2(H) = m_1(H) \sum_{F \supseteq H} \frac{m_2(F)}{b_1(F)} \quad (20)$$

and

$$m_1\|\|m_2(H) = m_1(H) \sum_{F \cap H \neq \emptyset} \frac{m_2(F)}{1 - b_1(\overline{F})}, \quad (21)$$

where $\mathcal{E}_i = \{H \subseteq X : m_i(H) > 0\}$, $i = 1, 2$. In this form, these rules appear quite opaque. If we sum these expressions over all subsets $H \subseteq A$ (which, strangely, Ichihashi and Tanaka do not do) we get the much clearer rules

$$b_1|b_2(A)$$
$$= \sum_{F \in \mathcal{E}_2} m_2(F) \frac{b_1(A \cup \overline{F}) - b_1(\overline{F})}{1 - b_1(\overline{F})} \quad (22)$$
$$= \sum_{F \in \mathcal{E}_2} m_2(F) b_1^{(d)}(A|F),$$

$$b_1\|b_2(A)$$
$$= \sum_{F \in \mathcal{E}_2} m_2(F) \frac{b_1(A \cap F)}{b_1(F)} \quad (23)$$
$$= \sum_{F \in \mathcal{E}_2} m_2(F) b_1^{(g)}(A|F),$$

and

$$b_1\|\|b_2(A)$$
$$= \sum_{F \in \mathcal{E}_2} m_2(F) \frac{b_1(A) - b_1(A \cap \overline{F})}{1 - b_1(\overline{F})} \quad (24)$$
$$:= \sum_{F \in \mathcal{E}_2} m_2(F) b_1^{(it)}(A|F)$$

It is immediately clear from these formulas that (22), (23), and (24) define belief functions, and that they each yield (8) when $b_1 = p$ and $b_2 = b$, and so extend not merely Jeffrey's rule (as Ishihaski and Tanaka note) but generalized probability kinematics as well, at least formally. It is also easy to check that $b_1|b_2(A) \geq b_2(A)$ and $b_1\|b_2(A) \geq b_2(A)$, which is in the spirit of generalized kinematics. It is not clear that this is true for $b_1\|\|b_2(A)$, due to a rather odd feature of what we have denoted $b_1^{(it)}(A|F)$, namely that it is possible that $b_1^{(it)}(F|F) < 1$! From our perspective this undermines (24) even as a formal generalization of Jeffrey's rule.

As for (22) and (23), they are subject to the same criticism that we have directed at other attempts to generalize Jeffrey's rule, in that they are furnished with no criterion which warrants their use. We are thus reminded once again that any revision rule deserving to be viewed as a generalization of conditionalization needs to be grounded both on new evidence (as incorporated in specified values or lower bounds on the values of the posterior) *and* on a judgement about the continued relevance of our prior uncertainty assessments (as incorporated in a conservation-of-conditional-probability criterion)

### Acknowledgements

Research supported by the National Science Foundation (DIR-8921269) and the Electric Power Research